%% file: cvpr_v2.tex
\documentclass[10pt,twocolumn,letterpaper]{article}

\usepackage{cvpr}
\usepackage{times}
\usepackage{epsfig}
\makeatletter
\@namedef{ver@everyshi.sty}{}
\makeatother
\usepackage{tikz}
\usepackage{pgfplots}
\usepackage{graphicx}
\usepackage{amsmath}
\usepackage{amssymb}
\usepackage{caption}
\usepackage{subcaption}
\usepackage{fancyhdr} 
\usepackage[toc,page]{appendix}
\usepackage{tabularx}
\usepackage{booktabs}
\usepackage{multirow}
\usepackage{ulem}
 \usepackage[export]{adjustbox}
 \usepackage{algorithm}
 \usepackage{mathtools}
 \usepackage{algorithmic}
 \usepackage{wrapfig,booktabs}
 \usepackage{wrapfig}
 \usepackage{array}
\usepackage{amssymb}
\usepackage{pifont}
\usepackage{cvpr}
\usepackage{times}
\usepackage{epsfig}
\usepackage{graphicx}
\usepackage{amsmath}
\usepackage{amssymb}
\usepackage{authblk}
\usepackage{enumitem}

\usepackage{tikz}

\usepackage[pagebackref=true,breaklinks=true,letterpaper=true,colorlinks,bookmarks=false]{hyperref}

\cvprfinalcopy 

\begin{document}
\title{DISCO: Dynamic and Invariant Sensitive Channel Obfuscation for deep neural networks}

\author{Abhishek Singh$^{1}$, Ayush Chopra$^{1}$,  Ethan Garza$^{1}$, Emily Zhang$^{1}$, Praneeth Vepakomma$^{1}$, 
\newline 
Vivek Sharma$^{1,2}$, Ramesh Raskar$^{1}$ \\
\small{$^{1}$ Massachusetts Institute of Technology,
$^{2}$ Harvard Medical School}
}

\maketitle

\begin{abstract}
Recent deep learning models have shown remarkable performance in image classification. While these deep learning systems are getting closer to practical deployment, the common assumption made about data is that it does not carry any sensitive information. This assumption may not hold for many practical cases, especially in the domain where an individual's personal information is involved, like healthcare and facial recognition systems.
We posit that selectively removing features in this latent space can  protect  the  sensitive  information  and  provide  better privacy-utility trade-off. Consequently, we propose DISCO which learns  a  dynamic and data driven pruning  filter to selectively obfuscate sensitive information in the feature  space.
We propose diverse attack schemes for sensitive inputs \& attributes and demonstrate the effectiveness of DISCO against state-of-the-art methods through quantitative and qualitative evaluation. Finally, we also release an evaluation benchmark dataset of 1 million sensitive representations to encourage rigorous exploration of novel attack and defense schemes at \url{https://github.com/splitlearning/InferenceBenchmark}.
\end{abstract}

\input{tex/intro}
\input{tex/rw}

\input{tex/method}
\input{tex/expts_new}
\input{tex/dataset_release}
\input{tex/results}
\input{tex/discussion}
\input{tex/conclusion}

{\small
\bibliographystyle{unsrt}
\bibliography{egbib}
}
\end{document}

%% file: tex/intro.tex
\section{Introduction}
Large deep neural network have resulted in breakthroughs across computer vision~\cite{yan2015deep}, speech recognition~\cite{amodei2016deep} and reinforcement learning~\cite{arulkumaran2017deep} with their success largely attributed to their ability to efficiently learn complex patterns from data.
The deployment of these algorithms in critical application domains such as healthcare and face-recognition has motivated a research focus on learning censored, unbiased and fair data representations to mitigate misuse by adversarial agents. Alternately, there can also be \textit{sensitive} information in data which the user would like to keep private but the learned representations may inadvertently encode. This sensitive information may manifest as sensitive inputs or attributes. Consider a setup where citizens consent to usage of face recognition in public spaces for identifying criminals. During inference, feature representations are extracted for faces and identification is performed by matching in the feature space over an indexed database. While this may be a well-intended initiative, a malicious adversary may seek to intercept the feature representations to i) reconstruct the input face image or ii) extract personal attributes such as race, age, gender etc. The citizens did not consent to sharing this sensitive information which could be used to compromise their privacy and in a way that is biased or unfair to them.  Exploring methods of improving privacy of the sensitive information (image, race, age, gender etc.) while preserving utility (identifying criminals) is the focus of this work.

Conventionally, research in privacy-aware machine learning has primarily focused on protecting training data from membership inference~\cite{shokri2017membership} and model inversion attacks~\cite{fredrikson2015model}, when i) training data is distributed over clients and ii) computation of training the model is out-sourced. For the former, distributed learning techniques such as federated learning~\cite{kairouz2019advances,konevcny2016federated} and split learning~\cite{splitlearning1,vepakomma2018split} are used, where clients communicate with a centralized server using weights and activations and the latter relies on homomorphic encryption~\cite{gentry2009fully,brakerski2014leveled} and secure enclaves~\cite{zhang2020hpress,ferraiuolo2017komodo}. Additionally, techniques such as multi-party computation~\cite{prabhakaran2013secure,evans2017pragmatic} and differential privacy~\cite{dwork2008differential,dwork2010differential,dwork2014algorithmic,cheu2019distributed} have been employed to improve the privacy in federated-learning. While effective for training, scaling these methods for deployment at inference is a challenge for a variety of reasons. First, in several cases computational limitations and intellectual property considerations limit keeping the entire model on a client device. Secondly, cryptographic methods for training deep networks ~\cite{hesamifard2017cryptodl,juvekar2018gazelle,nandakumar2019towards} are computationally very expensive operations which makes deploying models on the server infeasible when working with sensitive data. We posit that \textbf{collaborative inference}, where the inference network is distributed between client devices (client network) and a server (server network) which communicate via the \textit{split activations}, presents a viable alternative. While amenable to scalability, it is important to encode explicit measures of security in the intermediate activations to protect privacy of the sensitive inputs and attributes.

While not motivating private collaborative inference, a few recent works ~\cite{ li2019deepobfuscator, Roy_2019_CVPR} have attempted the related problem of attribute leakage~\cite{Roy_2019_CVPR,elazar2018adversarial,martinsson2020adversarial,bertran2019adversarially,sadeghi2019global} by focusing on adversarial representation learning (ARL). This couples together two entities, i) an adversarial network that seeks to extract a sensitive attribute from a given activation and, ii) a predictor network that intends to extract compact activations for accurate prediction of a task attribute (utility) while preventing the adversary from leaking the sensitive attribute (privacy). To balance this privacy-utility, ~\cite{Roy_2019_CVPR} designed an objective to maximize entropy of the adversary network and ~\cite{learnN2learn, li2019deepobfuscator} to minimize likelihood of the predictor on the sensitive attributes.

Motivated by the above observations, in this work, we first examine existing ARL methods which reveals the presence of high redundancy in learned representations. We posit that selectively removing features in this latent space can protect the sensitive information and provide better privacy-utility trade-off. Consequently, we propose DISCO which learns a dynamic and data driven pruning filter to selectively to obfuscate sensitive information in the feature space. We validate DISCO and other baseline with multiple attacks on inputs and attributes. We observe that DISCO consistently achieves superior performance by disentangling representation learning from privacy using the pruning filter.

To this end, the contributions of this work can be summarized as follows:
\begin{itemize}[nosep]
    \item We introduce DISCO, a dynamic scheme for obfuscation of sensitive channels to protect sensitive information in collaborative inference. DISCO provides a steerable and transferable privacy-utility trade-off at inference.
    \item We propose diverse attack schemes for sensitive inputs and attributes and achieve significant performance gain over existing state-of-the-art methods across multiple datasets.
    \item To encourage rigorous exploration of attack schemes for private collaborative inference, we release a benchmark dataset of \textbf{1 million} sensitive representations.
\end{itemize}

%% file: tex/rw.tex
\section{Related Work}
\begin{figure*}[t!]
\centering
\vspace{-8mm}
{\includegraphics[width=1.8\columnwidth]{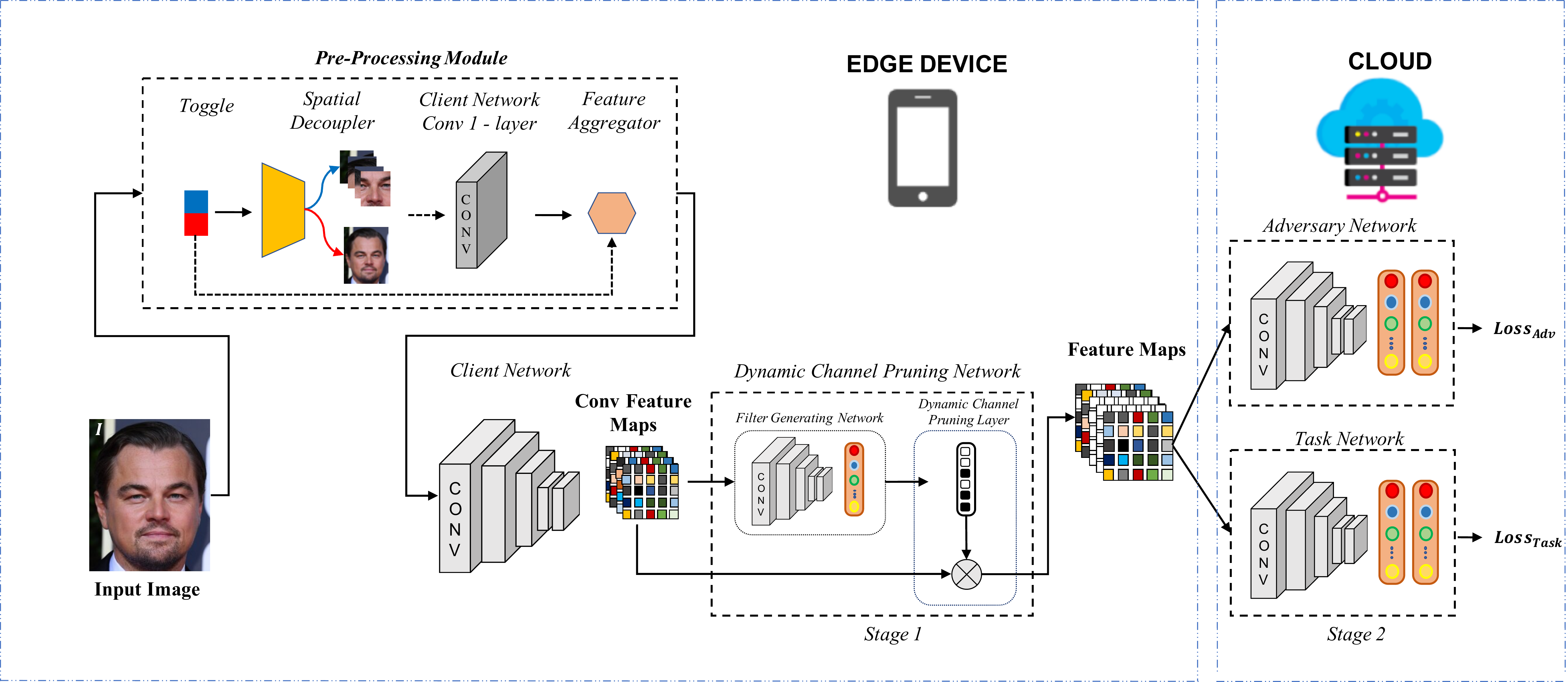}} 
\caption{\textbf{DISCO for Privacy}. Input to the network is an image, as well as task labels and attribute labels to hide. The network is jointly optimized with a task
objective to adaptively hide a given attribute without drop in performance of the target task.}
\label{fig:network-arch}
\vspace{-5mm}
\end{figure*}
\textbf{Private Representation Learning}  \cite{konevcny2016federated,splitlearning1} propose mechanisms which allow for learning on data distributed across multiple agents with raw training data never leaving the corresponding client device. \cite{abadi2016deep} further improves \cite{konevcny2016federated} by adding differentially private noise to weights of the trained model to prevent reconstruction of training data by inversion attacks. That said, techniques such as \cite{abadi2016deep} are largely optimized to protect training data. In contrast, there is limited research on methods for privacy during inference via privatized activations. Majority of the works in private inference use ARL~\cite{Roy_2019_CVPR,sadeghi2019global,learnN2learn,bertran2019adversarially,li2019deepobfuscator,vepakomma2020nopeek} to learn a feature extractor that minimizes sensitive information leakage. Bertran et al.~\cite{bertran2019adversarially} apply transformation in the image space to ensure server's input remains an image. \cite{vepakomma2020nopeek} introduced a distance correlation based regularization to decouple intermediate activations from input data while preserving performance on task attribute. While efficacy of these methods depend upon the convergence of min-max optimization, our work separates the feature extraction and privatization module giving guaranteed reduction in mutual information. In this work, we explore methods that seek to reduce redundancy and semantic integrity of activations to mitigate attacks on sensitive information.

\textbf{Natural Pre-Image} is a class of diagnostic techniques which are designed to reconstruct input image from intermediate activation and find utilization in computer vision tasks such as denoising, super-resolution etc. \cite{DIP} leverages a randomly-initialized neural network and a hand crafted prior to invert deep neural representations and reconstruct the input. \cite{dosovitskiy2016inverting} seeks to train a decoder offline to learn to predict the input distribution. We leverage expected pre-image methods to formalize diverse attack schemes on sensitive inputs.

\textbf{Bias in Machine Learning} is a recent direction of ML research focused on two key problems: identifying and quantifying bias in datasets, and mitigating its harmful effects. The bias routinely manifests as some attributes of the input (eg. age, race, gender for faces). A popular category of techniques involve adversarial representation learning ~\cite{wang2020towards,learnN2learn, alvi2018turning} to mitigate the impact of the bias attribute on the task attribute. This family of adversarial mitigation techniques aligns with this work on selective privacy, with the private attribute analogous to the bias attribute, and a corresponding state-of-the-art ~\cite{learnN2learn} forms one baseline for our study.

\textbf{Part-based Representation Learning} involves splitting the image into several stripes to learn local representations and has achieved promising performance on computer vision tasks such as person re-identification which involves image retrieval under occlusions and partial observability. While sophisticated learning based partitioning methods have been explored~\cite{reid-1, reid-2, reid-3, reid-4}, methods such as ~\cite{mgn} have achieved outstanding performance with trivial deterministic splitting. In this work, we adapt the static part-based techniques to decouple the intra-channel semantic consistency of convolutional activations for improving privacy-utility trade-offs in collaborative inference. 

\textbf{Channel Pruning} is a prevalent technique for deep network compression to minimize computational complexity and accelerate inference \cite{cheng2017survey}. While most methods interleave pruning with the training phase \cite{grad-cha-prun-bmvc, molchanov2019taylor, NIPS2018_7367}, there has been recent focus on pruning at inference \cite{gao2018dynamic}. \cite{grad-cha-prun-bmvc} gradually prunes channels at fixed intervals during training using a feature relevance score to minimize compute cost. \cite{gao2018dynamic} propose dynamic feature boosting and suppression (FBS) to predictively amplify salient convolutional channels and skip unimportant ones at run-time for accelerated inference. In this work, our proposed method can be aligned with channel pruning but optimizes for a different objective of preventing leakage of sensitive information. 

\textbf{Filter Generating Networks~(FGN)} ~\cite{dfn,stn}, there is very limited literature on FGNs. One such module, the ``Spatial Transformer" network, is proposed by Jaderberg et al.~\cite{stn}. This spatial transformer module applies an affine transformation to feature maps to do translation and rotation for improved classification. Following~\cite{stn}, all these recent works~\cite{dfn,dcl,dyncnn} utilize the same concept to learn a steerable filter~\cite{dfn}, weather prediction filter~\cite{dcl}, an image enhancement filter~\cite{dyncnn}, and a dynamic motion motion representation filter~\cite{dynamo} using source-target image pairs.  In contrast to these works, our focus is to learn dynamic filters that selectively prune channels which leak sensitive attributes without dropping in performance on the target task. The output of our dynamic channel pruning filters are binary (0 or 1) in nature, where 0 masks~(or deactivates) channels that contribute to sensitive attributes, and 1 unmasks channels that contribute to the target task at hand.

%% file: tex/method.tex
\begin{figure}
    \centering
    \includegraphics[width=.47\textwidth]{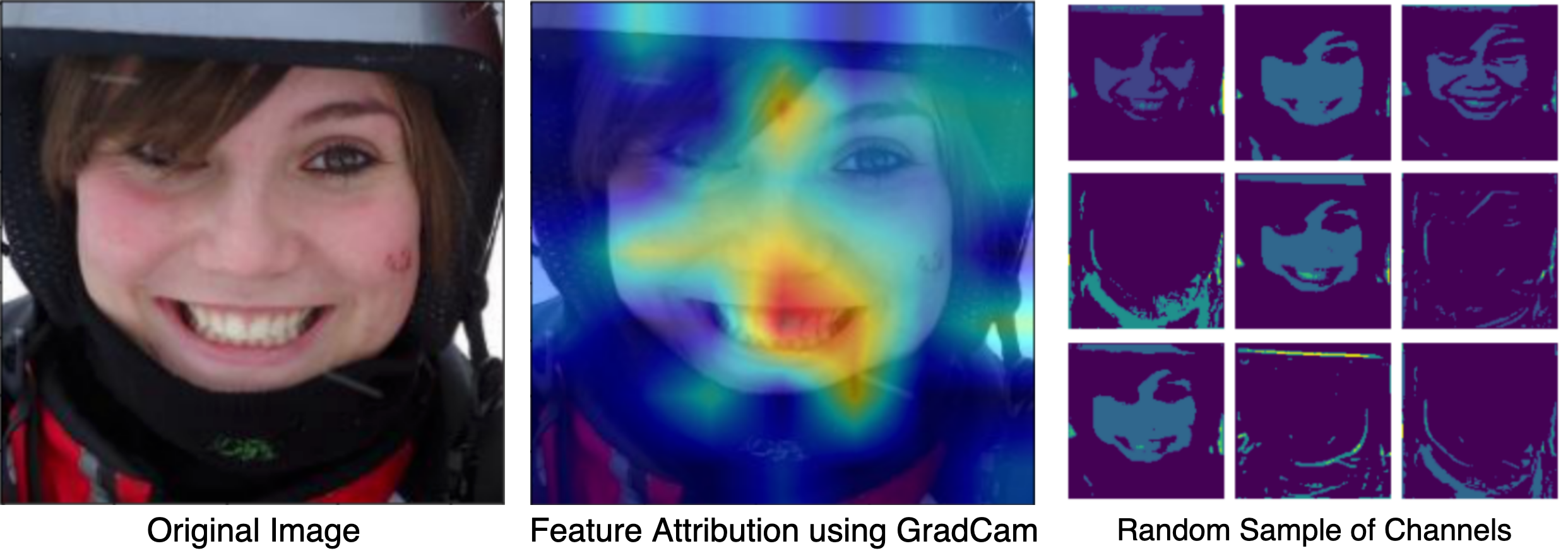}
    \caption{a) Input Image and Grad-CAM visualization from ResNet-18 classifier b) Corresponding convolution representations which encode inter-channel redundancy and preserve intra-channel semantic integrity.\label{fig:premise-validation}}
    \vspace{-0.3cm}
\end{figure}

\section{Methodology}

First, we introduce the attack and threat models and then define the privacy considerations for our work. Finally, we formalize our privacy evaluation setup and delineate our proposed method DISCO: \textit{Dynamic and Invariant Sensitive Channel Obfuscation} for protecting sensitive information in latent representation.

\subsection{Formulation}
\label{sec:attack}
\paragraph{Setup.}
Consider a parameterized model $f(\theta; \cdot)$ trained to estimate the target attribute $y \in \mathcal{Y}_1 $ for a given input image  $x \in \mathcal{X} $. In many scenarios, $x$ may be a sensitive input or have a sensitive attribute $\hat{y} \in \mathcal{\hat{Y}} $. Considerations for balancing compute feasibility and privacy has motivated private collaborative inference schemes \cite{Roy_2019_CVPR, li2019deepobfuscator} that split $f(\theta; \cdot)$ into $f_1(\theta_1; \cdot)$ and $f_{2}(\theta_2; \cdot)$ where:

$$f_{1}(\theta_1; x) \in F_1:\mathcal{X}\times\Theta_1\rightarrow \mathcal{Z}$$
$$f_2(\theta_2;z) \in F_2:\mathcal{Z}\times\Theta_2\rightarrow \mathcal{Y}_1$$

such that $f_2=(\theta_2;f_1(\theta_1; x))$ and $\theta=\mathbf{\{}\theta_1, \theta_2\mathbf{\}}$. We refer to this as as \textit{traditional} setup for collaborative inference. We formalize $f_1$ as the \textit{client network} that is executed on a trusted device and $f_{2}$ as a \textit{task network} which executes on an untrusted server using the \textit{client activation} $z = f_{1}(\theta_1; x)$.

\textbf{Threat Model.}
Under our threat model, the untrusted server could attempt to learn sensitive information about $x$ by inferring an arbitrary sensitive attribute $\hat{y}$ or by reconstructing $x$ itself. As a concrete example, $x$ may be a face image with $y$ as gender and $\hat{y}$ as racial identity.
For the evaluation and algorithm design purposes, we build a proxy adversary that attempts to approximate the real world adversary. This proxy adversary is parameterized with an \textit{adversarial network} $f_{3}(\theta_{3}; \cdot)$ that may intercept the payload $z$ to extract the sensitive input $x$ or the attribute $\hat{y}$.
\begin{figure*}
    \centering
    \includegraphics[width=\textwidth]{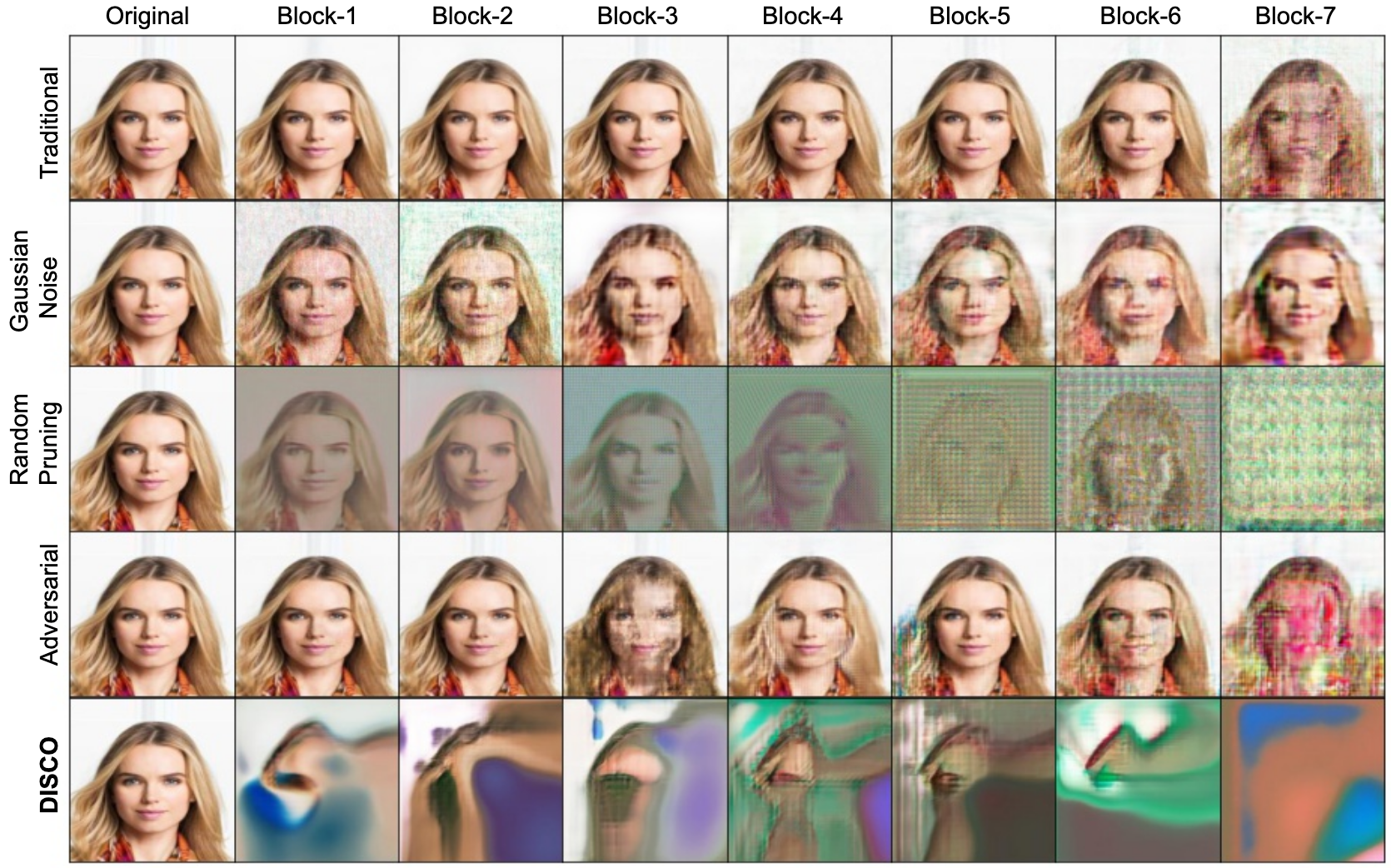}
    \caption{\textbf{Reconstruction results on CelebA~\cite{liu2018large}}: All of the reconstructed images are obtained from the activations using the likelihood maximization attack. We generate activations from the ResNet-18~\cite{resnet} architecture where a set of convolution, batch normalization, and activation layers are grouped under a block. The first column shows the original sensitive input and remaining columns show its reconstruction across different blocks. For gaussian noise we use $\mu=-1.,\sigma=400$, this is the amount of noise at which the learning network gets utility close down to random chance. \textit{Adversarial} refers to the set of techniques for filtering sensitive information using adversarial learning~\cite{li2019deepobfuscator,learnN2learn}. For \textit{DISCO} and \textit{Random Pruning} we use a pruning ratio of $R=0.6$. \label{fig:reconstruction_fairface}}
    \vspace{-0.5cm}
\end{figure*}
\textbf{Attack Model.}
The adversary may utilize the activation $z$ to perform a \textit{reconstruction attack} to recover the sensitive input or a \textit{leakage attack} to extract the sensitive attribute. We define the following attack models for the sensitive information $z$:
\begin{itemize}
    \item \underline{\textit{Supervised Decoder:}} In this attack setting, the adversary leverages a small number of $(z, \hat{y})$ pairs to train a neural network $\hat{f}(\hat{\theta})$ such that $\hat{y}=\hat{f}(z)$. The practical validity of this attack is in the scenarios where some finite number of pairs $(z, \hat{y})$ is obtained through a malicious or colluding client who is also participating in the collaborative inference setting. This attack scheme is inspired from~\cite{dosovitskiy2016inverting,DBLP:journals/corr/MahendranV14,DBLP:journals/corr/DosovitskiyB15}. Another practical scenario for this attack is where the pairs $(x, \hat{y})$ from a similar distribution is publicly available, in such a case the client can train an auto-encoder and use the trained decoder for the attack. This technique can utilized for both \textit{reconstruction attack} and \textit{leakage attack}.
    \item \underline{\textit{Likelihood Maximization:}} Unlike the above scheme, here $(z, \hat{y})$ pairs are not needed to reconstruct the sensitive input, instead, the attacker uses the weights $\theta_1$ of the \textit{client network} and a randomly initialized network $\hat{f}(\hat{\theta};\cdot)$ that generates an image $\hat{x}$ to produce $\hat{z} = f_1(\theta_1, \hat{x})$. Then the loss $\ell_2(\hat{z}, z)$ between random and sensitive activation is minimized by optimizing $\hat{\theta}$. This attack scheme is inspired by the deep image prior~\cite{DIP} for feature inversion. This attack is only applicable to the sensitive input protection and not sensitive attribute. This attack setting is stronger and harder to defend against because it does not require access to the $(z,\hat{y})$ pairs.
\end{itemize}

\textbf{Privacy.}
Following the setup described in Hamm~\textit{et al.}~\cite{JMLR:v18:16-501}, we measure privacy as the expected loss over the estimation of sensitive information by the adversary. This privacy loss $L_{priv}$, given $\ell_p$ norm, for an adversary can be stated as:
$$L_{priv}(\cdot)\triangleq E[\ell_p(\hat{f}(z), \hat{y})]$$
Under this definition, releasing sensitive information while preserving privacy manifests as a min-max optimization between the data owner and the attacker. For training the model parameters, we use a proxy adversary from which gradients can be propagated. We formalise our setup as an analogue but relax the non-invertibility assumption made by Hamm~\textit{et al.}~\cite{JMLR:v18:16-501} for the client $f_{1}$, following \cite{DBLP:journals/corr/AroraLM15}, to generalize the attack surface to sensitive inputs. Additional details for the privacy framework are included in the \textit{supplementary}.

\subsection{Premise Validation}
Adversarial representation learning (ARL) is the existing state-of-the-art approach for performing private inference ~\cite{Roy_2019_CVPR, li2019deepobfuscator, learnN2learn} on sensitive data. Consider Figure~\ref{fig:premise-validation} which visualizes the face image and the learned client activation in ARL~\cite{learnN2learn}. We note the following observations: a) the learned activations have high inter-channel redundancy, and b) individual feature maps preserve semantic integrity of the input image, especially with shallower client networks. Since gradient attribution in convnets is spatially localized~\cite{grad-cam}, we posit that reducing this inter-channel redundancy and perturbing the intra-channel integrity of \textit{client activations} can help achieve better privacy-utility trade-offs.

\subsection{DISCO}
DISCO, depicted in Figure~\ref{fig:network-arch}, is composed of three key entities: a client, a predictor, and an adversary. The client transforms the input image to generate \textit{client activations} which are communicated to the predictor for inferring the task attribute but can be collected by an adversary.

\textbf{a) Client} owns the sensitive information. Given an input $x \in R^{3\times H\times W}$, this entity participates in the collaborative inference and intends to achieve privacy in the \textit{client activations} $z$ it communicates. 

Initially, $x$ is passed through the \textit{pre-processing module} where the \textit{spatial decoupler} first decomposes it into $d^2$ disjoint spatial partitions $P_{i} \in R^{3\times\hat{H}\times\hat{W}}$ for $i = \{0, 1, 2, ...., d^2\}$ with $\hat{H}=H/d, \hat{W}=W/d$. Next, each of the partitions $P_{i}$ is resized back to $H\times W$ and passed through a convolutional layer (with $F$ filters) to generate $\hat{P}_{i} \in R^{F\times H'\times W'}$. Finally, the \textit{feature aggregator} generates an aggregated representation $A \in R^{d^2\times H'\times W'}$ by averaging across channels and re-stacking each $\hat{P_{i}}$. $A$ is then communicated to the client network. Here, we note that $d^2 = F$ in our pre-processing module so that the spatial decoupler can be easily bypassed (toggled-off) without altering the rest of the network architecture. The underlying idea to use \textit{pre-processing module} is to ensure pruning of channels in $z$ leads to removal of unique spatial information. If not performed, the redundancy present across channels in $z$ would allow an attacker to recover the full image even from a pruned $z$.

Next, the \textit{client network} takes $A$ as input and generates an intermediate activation $\hat{z} \in R^{C''\times H''\times W''}$. Finally, the \textit{filter generating network} $g(\phi,\hat{z})$ parameterized by $\phi$ takes $\hat{z}$ as input and generates a feature map score $F \in R^{C''}$ for each channel in $\hat{z}$. The $F$ channel pruning filters are weakly discretized using sigmoid with temperature (to avoid introducing
discontinuity) and then thresholded to obtain a binary vector $b$. Then $b$ is multiplied channel wise with $\hat{z}$ to produce a pruned feature volume $z$, the \textit{client activation}, with channels leaking the sensitive information masked out (or deactivated) in the latent space. Note that $F$, the feature map score, is conditioned on $\hat{z}$
(hence $x$) and is thus generated dynamically on run-time per sample basis. A key idea of DISCO is to disentangle representation learning from privacy via the learned pruning filter. The hyper-parameter pruning ratio $R$ governs the number of active channels in the pruning filter and helps regulate the privacy-utility trade-off.

\textbf{b) Predictor} is an untrusted entity that receives the \textit{client activations} ($z$) and executes the \textit{task network} ($f_{2}$) to estimate the task attribute ($y$). The task network is optimized on the conventional loss function ($\ell_u$) used for the task. In this paper we consider image classification as taska and hence use cross entropy loss ($\ell_{cce}$).

\textbf{c) Adversary} also receives the client activations ($z$) and executes the \textit{adversarial network} ($f_{3}$) with the intent of extracting \textit{sensitive} information - input or attribute. The adversary performs reconstruction attacks for obtaining the sensitive inputs or attribute leakage attacks to infer sensitive attributes. During the training, we design a proxy adversary that has access to the sensitive inputs ($x$) and attributes ($\hat{y}$). For reconstruction attacks, the adversarial network is a decoder module optimized using $\ell_{1}$ loss against the input $x$. For attribute leakage attacks, the adversarial network is a convolutional classifier module optimized using $\ell_{cce}$ loss against the sensitive attribute $\hat{y}$. The adversary loss can be summarized as :

\[ l_{a} =  \begin{cases} 
      \ell_{1}(f_3(z), x) & mode =  SI \\
      \ell_{cce}(f_3(z), \hat{y})) & mode = SA \\
   \end{cases}
\]
where, mode $\in$ [SI, SA] represents attack on sensitive input (SI) or sensitive attribute (SA). Note that $f_3$ is a proxy adversary used for training purposes while $\hat{f}$ is the real world adversary which will be used for attack during evaluation.

\subsection{Training}
The utility of the task during inference depends upon parameters $\theta_1,\phi,\theta_2$ learned during the training stage and can be expressed as
\begin{equation}
\label{eq:util}
L_{util}(\theta_1,\phi,\theta_2)\triangleq E[\ell_u(f_2(g(f_1(x;\theta_1);\phi);\theta_2), y)]    
\end{equation}

As described previously we use a proxy adversary during the training and evaluation of our setup as described by the evaluation function $L_{priv}$ to train the pruning network.
\begin{equation}
\label{eq:priv}
L_{priv}(\theta_1,\phi,\theta_3)\triangleq E[\ell_a(f_3(g(f_1(x;\theta_1);\phi);\theta_3), \hat{y})]    
\end{equation}

$\theta_3$ is the parameters for the proxy adversary used during the training and evaluation. $\ell_u$ and $\ell_a$ is the loss function used for evaluating utility and privacy respectively. The adversary network and task network have access to supervised data and attempt to minimize their losses $L_{util}$ and $L_{priv}$ respectively. The filter generating network is trained to minimize $L_{util}$ and maximize $L_{priv}$, simulating an implicit min-max optimization for these two components. The client network parameters are only optimized to minimize $L_{util}$. We deliberately restrict $\theta_1$ for minimizing $L_{util}$ and not maximize $L_{priv}$ to ensure that the filter generating network generalizes and does not trivially utilize representations learned by the $\theta_1$. This makes our explicit privatizing module $g(\phi,\cdot)$ one of the big differentiating factor of our work from existing ARL based methods~\cite{li2019deepobfuscator,Roy_2019_CVPR,learnN2learn}. We posit that this facilitates the filter generating network to specialize at pruning by identifying the privacy leaking channels. This overall objective can be summarized as:
\begin{equation}
\label{eq:total_loss}
    \min_\phi \left[\max_{\theta_3} -L_{priv}(\theta_1, \phi, \theta_3) + \rho \min_{\theta_1,\theta_2} L_{util}(\theta_1, \phi, \theta_2)\right]
\end{equation}

Here, $\rho$ is chosen as a hyper-parameter to trade-off between accuracy and privacy.

\subsection{Prediction}
During the inference stage, computation for feature extraction $\hat{z}=f_1(x;\theta_1^*)$ and pruning $z=g(\hat{z};\phi^*;R)$ is performed on the trusted system and $z$ is sent to the untrusted party. The value of pruning ratio $R$ governs the total number of channels to be pruned from $z$ and allows adjusting for the privacy and utility trade-off during runtime.

\subsection{Generalization}
In the equation~\ref{eq:total_loss}, the goal is to perform a joint optimization where $\theta_1,\phi,\theta_2,\theta_3$ are the parameters of the \textit{filter generating network}, \textit{client network}, and \textit{server network} respectively. Let $\theta_1^*,\phi^*,\theta_2^*, \theta_3^*$ be the solution for the parameters we obtain by minimizing the expected loss. Let $\hat{\theta}_1,\hat{\phi},\hat{\theta}_2,\hat{\theta}_3$ refers to the empirical minimizer of the above mentioned joint optimization. As noted before, we adapt to the setup described by Hamm~\cite{JMLR:v18:16-501}. However, a significant difference lies in the fact that $\theta_1$ is not trained to minimize $L_{priv}$ as this is to improve generalization of the $\phi$ across a different set of $\theta_1$. The remaining parameters remain analogous to the min-max filters described in~\cite{JMLR:v18:16-501} .Following on that, we describe the joint loss as follows
$$L_J(\theta_1,\phi,\theta_2,\theta_3)=\rho L_{util}(\theta_1,\phi,\theta_2) - L_{priv}(\theta_1, \phi, \theta_3)$$
Let $D$ be the original unknown data distribution and $S$ be a set of samples obtained from the true distribution for calculating empirical loss then the empirical and expected loss can be bounded as follows, giving a generalization bound.
\begin{align*}
    |E_D(L_J(\theta_1^*,\phi^*,\theta_2^*, \theta_3^*)) - E_S(L_J(\hat{\theta}_1,\hat{\phi},\hat{\theta}_2,\hat{\theta}_3))|\leq\\
    2\sup_{\theta_1,\phi,\theta_2,\theta_3}\lvert E_D(L_J(\theta_1,\phi,\theta_2,\theta_3)) 
    - E_S(L_J(\theta_1,\phi,\theta_2,\theta_3))\rvert    
\end{align*}
For more details, we refer the reader to the proof of theorem 1 shown in~\cite{JMLR:v18:16-501}. The equation above gives the bound on generalization error.
\subsection{Effect of channel pruning on mutual information} We now study the effect of applying channel pruning of activations at the output of the client network with regards to the mutual information between the raw sample and the pruned activations. Inspired by the theoretical analysis in~\cite{DBLP:journals/corr/abs-1905-11814}, we extend and adapt it to our setup of analyzing the reduction in mutual information between the \textit{sensitive input} and \textit{client activations} upon performing random pruning. We use the superscript notation $f_1^{k}(\theta_1^k;x)$ to denote the output of $k$'th layer of client network. We compare this with regards to no pruning and random pruning at the $k$'th layer of the client network as shown below.\\ \textbf{Pre-pruning:} The negative of the mutual information between the raw data and the output of 1'st layer prior to applying the pruning is given by \begin{align*} 
-\mathcal{I}(x;f_1^{1}(\theta_1^1;x))&=-\mathcal{H}(f_1^1(\theta_1^1;x))-\mathcal{H}(f_1^1(\theta_1^1;x)|x)\\&=-\mathcal{H}(f_1^1(\theta_1;x))
\end{align*} as $-\mathcal{H}(f_1^1(\theta_1;x)|x)=0$, due to $f_1^1(\cdot)$ being a deterministic function.
Upon applying the data processing inequality, we have that the mutual information between the output of the $k$'th layer and the raw data satisfies: $$ \mathcal{I}(x;f_1^{k}(\theta_1^k;x))\leq \mathcal{I}(x;f_1^{k-1}(\theta_1^k-1;x))\leq\ldots \leq \mathcal{I}(x;f_1^{1}(\theta_1^1;x))$$ where, we have the following relation $\mathcal{I}(x;f_1^{k}(\theta_1^k;x))=\mathcal{H}(f_1^k(\theta_1^k;x))$.\\
\textbf{Post-pruning:}
The mutual information after random pruning can be represented as a multiplication of the outputs at the $k$'th layer with a Bernoulli random variable $\mathcal{P}$ as $\mathcal{I}(x;f_1^k(x,\theta_1^k).\mathcal{P})$.  In addition to the form of data processing inequality used in analysis of pre-pruning; there is an equivalent form of the classical data processing inequality given by  $$-\mathcal{I}(x;f_1^k(x,\theta_1^k).\mathcal{P}) \geq -\mathcal{I}(f_1^k(x,\theta_1^k);f_1^k(x,\theta_1^k).\mathcal{P})$$ Upon expanding this upper bound using entropy terms we get $$\mathcal{I}(x;f_1^k(x,\theta_1^k).\mathcal{P}) \leq \mathcal{H}(f_1^k(\theta_1^k;x)) - \mathcal{H}(f_1^k(\theta_1^k;x)|f_1^k(\theta_1^k;x).\mathcal{P})$$ But $\mathcal{H}(f_1^k(\theta_1^k;x))$ is the mutual information in the case of pre-pruning as analyzed above. Therefore the decrease in information about raw data post-pruning is given by the term $\mathcal{H}(f_1^k(\theta_1^k;x)|f_1^k(\theta_1^k;x).\mathcal{P})$.  Upon applying the Bayes rule (for conditional entropy), this term exactly equals: $$ \mathcal{H}(f_1^k(\theta_1^k;x).\mathcal{P}|f_1^k(\theta_1^k;x)) + \mathcal{H}(f_1^k(\theta_1^k;x))-\mathcal{H}(f_1^k(\theta_1^k;x).\mathcal{P})$$ Since the term $f_1^k(\theta_1^k;x)$ is independent of the noise $\mathcal{P}$, the above can be further rearranged as $$ \mathcal{H}(f_1^k(\theta_1^k;x).\mathcal{P}|f_1^k(\theta_1^k;x)) + \mathcal{H}(f_1^k(\theta_1^k;x))-\mathcal{H}(f_1^k(\theta_1^k;x))-\mathcal{H}(\mathcal{P})$$ which simplifies to $\mathcal{H}(f_1^k(\theta_1^k;x).\mathcal{P}|f_1^k(\theta_1^k;x)) - \mathcal{H}(\mathcal{P})$. As we chose $\mathcal{H}(\mathcal{P})$ to be a Bernoulli random variable; upon considering its success probability to be $p$ (lower-case) and probability of failure to be $q=1-p$, we have -$\mathcal{H}(\mathcal{P})=p log(p) + q log(q)$. Therefore, upon performing random pruning the decrease in mutual information amounts to $$\mathcal{H}(f_1^k(\theta_1^k;x).\mathcal{P}|f_1^k(\theta_1^k;x)) - \mathcal{H}(f_1^k(\theta_1^k;x)) +p log(p) + q log(q) $$ while the mutual information post-pruning is upper bounded by $\mathcal{H}(f_1^k(\theta_1^k;x).\mathcal{P}|f_1^k(\theta_1^k;x)) +p log(p) + q log(q) $.

\section{Discussion: Dynamic Design of \textit{DISCO}}
A key idea behind DISCO is the decoupling of privacy considerations from representation learning using the dynamic pruning filter. We analyse the dynamic formulation of this design along the following dimensions:
\begin{itemize}[nosep]
    \item \underline{Dynamic Private Representations}: The filter generating network in DISCO estimates the pruning filter for each input, independently at run-time. Since different convolutional filters are known to activate differently~\cite{gao2018dynamic}, the dynamic channel pruning in DISCO enables more personalized identification of sensitive channels for each input resulting in better privacy-utility trade-offs.
    \item \underline{Dynamic Integration}: We train DISCO in two phases as i) train the client and the predictor networks to maximize utility ii) train filter generating network with predictor and the (proxy) adversary to minimize privacy leakage and preserve utility. Decoupling of $g$ from $f_1$ enables private \textit{expert filters} that can obfuscate sensitive attributes and be employed by a network running DISCO. For example, one can build a dictionary of DISCO modules for different sensitive attributes for faces such as race, gender, eyeglasses, and etc. can be trained and used by different vendors based on their context for privacy and utility.
    \item \underline{Dynamic Privacy Utility Trade-offs}: All previous methods weight seek to balance privacy-utility during training by weighting the corresponding losses. However, once the model is trained, the privacy-utility trade-off is frozen. In contrast, DISCO can allow dynamically varying privacy-utility at inference by tweaking the pruning ratio ($R$). However, this would also require the server's parameters ($\theta_2$) to be trained with different $R$. This dynamic adjustment enables one to continuously control the privacy offered by deployed systems without having to interrupt or retrain the machine learning service from scratch.
\end{itemize}

%% file: tex/expts_new.tex
\section{Experiments}
\paragraph{Datasets} We conduct experiments with the following datasets:
\begin{itemize}[nosep]
    \item Fairface \cite{karkkainen2019fairface} dataset consists of 108,501 images, with race, gender, and age groups. The dataset is designed with the emphasis of balanced race composition which we preserve in our experimental train and test sets. For our experiments, the task attribute is gender and the sensitive attribute is race.
    \item CelebA \cite{liu2018large} consists of 202,599 celebrity face images across 10,177 identities, each with 40 attribute annotations. For our experiments, we define the task attribute as emotion and sensitive attribute as gender.
    \item CIFAR \cite{krizhevsky2010convolutional} consists of 60000 32x32 colour images in 10 classes, with 6000 images per class. There are 50000 training images and 10000 test images. We manually label each of the 10 classes as living or non-living. For our experiments, the task attribute is the class label and sensitive attribute is living/non-living, as introduced in ~\cite{Roy_2019_CVPR}.
\end{itemize}

\begin{figure}
    \centering
    \includegraphics[width=.8\columnwidth]{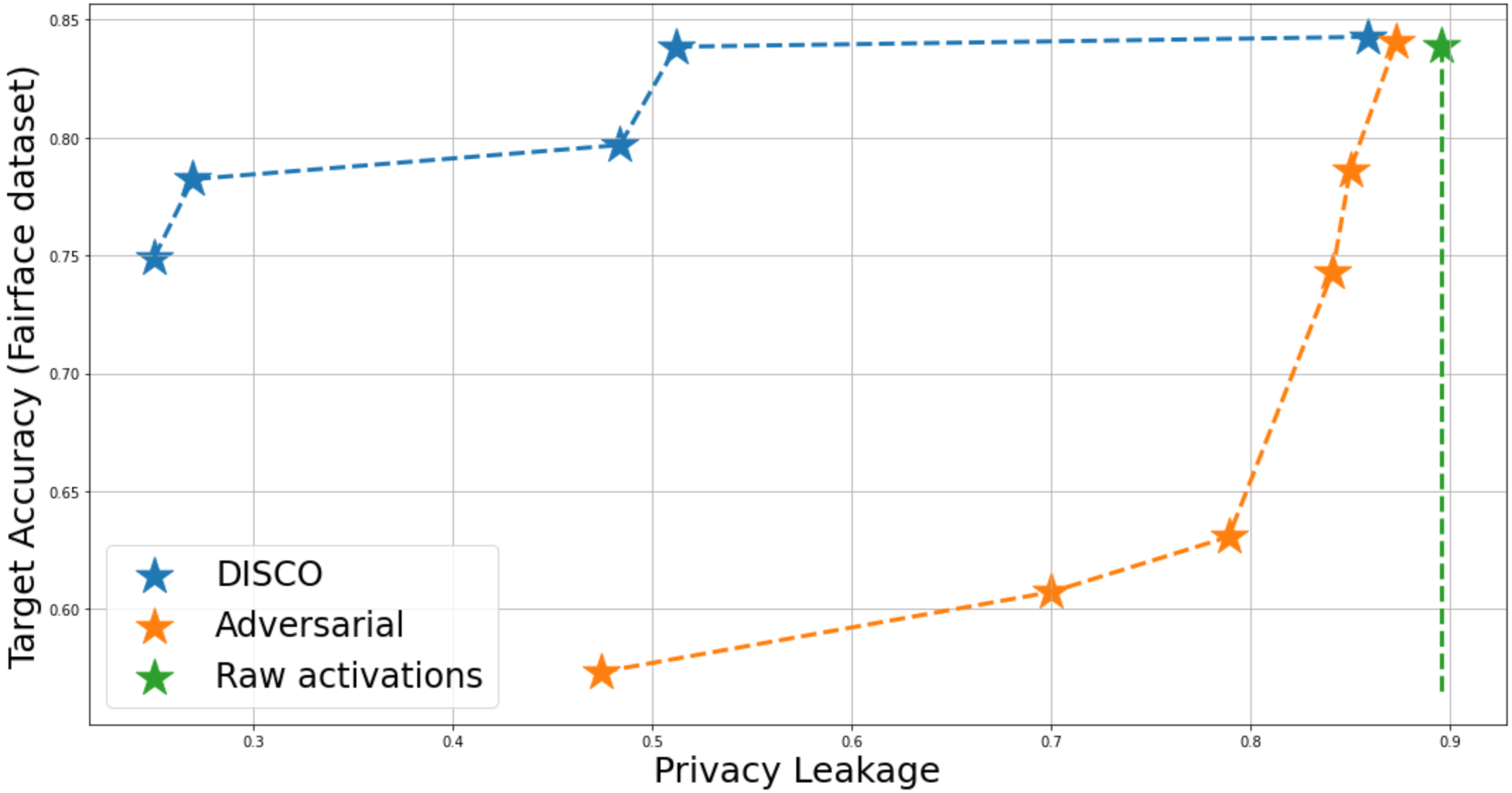}
    \caption{\textbf{Privacy-Utility Trade-off}: We vary the pruning ratio $R$ for DISCO and $\lambda$ trade-off parameter for ARL~\cite{li2019deepobfuscator,learnN2learn}. Leakage is measured as SSIM score between inputs and reconstruction.}
    \label{fig:tradeoff}
    \vspace{-6mm}
\end{figure}

\paragraph{Implementation Details} Experiments are implemented using Pytorch and conducted using NVIDIA Tesla V100 GPUs. The backbone network is ResNet-18~\cite{resnet} with the \textit{client activations} obtained from the block-4, unless specified otherwise.\\
\textbf{Hyper-parameters and Experimental Setup:} All of the experimental setup is implemented in PyTorch and we will be releasing the codebase for all of different quantitative and qualitative experiments, with the random seeds used in all of the experiments.\\
\textbf{Network architecture}: We describe four distinct networks in the section 3, \textit{client network}, \textit{filter generating network}, \textit{adversary network}, \textit{task network}. We use ResNet-18~\cite{resnet} as the base architecture for all of the four networks. For alignment of the architecture we experiment with the different blocks of the ResNet architecture and split the network such that output of the \textit{client network} is fed to all three \textit{filter generating network}, \textit{adversary network}, and \textit{task network}. The \textit{filter generating network} has same number of neurons in the final fully connected layer as number of channels in the output produced by \textit{client network}. The sigmoid temperature is 0.03 for the filter generating network. We adapt the ResNet backbone for \textit{adversary network} when the protected attribute is sensitive input since it requires to build a generative model conditioned on \textit{client activations}. We use a transpose convolution based architecture that upsamples the feature map to a higher dimensionality resulting in final image.\\
\textbf{Pre-processing module} described in the section 3.3.a is composed of a single convolution layer and a \textit{spatial decoupler} that splits the feature-map into $d^2$ spatially disjoint partitions. For an image size of $112$ and target $d^2$ to be $64$, the resulting featuremap size is $14\times14$ that gets rescaled back to $112\times112$ using bilinear interpolation. We keep the value of the $d^2$ as $64$ to make sure that the averaging in the channel space results in $64$ distinct feature maps that can be fed into the remaining of the architecture, this allows compatibility of the \textit{pre-processing module} with off the shelf architectures.\\
\textbf{Optimizer}: We use SGD optimizer with momentum~\cite{qian1999momentum} for all of the networks with a learning rate of $0.01$

\textbf{Evaluation Metrics} We measure utility using top-1 accuracy on the task attribute. For attacks on sensitive inputs, we measure privacy using $\ell_1$ loss, SSIM and PSNR~\cite{hore2010image} between reconstructed and input image and top-1 accuracy on the private attribute for attacks on sensitive attributes.

\textbf{Baselines}
For attacks on sensitive attributes, we baseline with ARL based methods~\cite{learnN2learn, Roy_2019_CVPR, li2019deepobfuscator} which are state-of-the-art on attribute leakage. For attacks on sensitive inputs, we baseline with ~\cite{learnN2learn, li2019deepobfuscator} and two randomized variants of DISCO where we perform: i) random pruning ii) gaussian noise. Finally, for both sensitive inputs and attributes, we also compare with a traditional CNN model, denoted as\textit{traditional}, with no activation privacy, this has been studied in Osia et al~\cite{osia2020hybrid}.
\begin{table*}[]
\tabcolsep=0.2cm
\centering
\begin{tabular}{l|c|c|c|c|c|c}
\toprule
\multirow{2}{*}{Method} & Privacy  & Utility  & Privacy & Utility & Privacy & Utility \\ 
 &  (Fairface) $\downarrow$ &  (Fairface) $\uparrow$ &  (CelebA) $\downarrow$ &  (CelebA) $\uparrow$ & (CIFAR10) $\downarrow$ &  (CIFAR10) $\uparrow$ \\ 
\midrule
\cite{osia2020hybrid}    & 0.319    & \textbf{0.824}   & 0.729   &  \bf{0.916}  &  0.912 & 0.498 \\ 
\midrule
DISCO    & \textbf{0.190}           & 0.815           & \textbf{0.612} & 0.910 & ~\textbf{0.223} & 0.9198 \\
\midrule
\cite{Roy_2019_CVPR}  & {0.236}      & {0.802}  & {0.780}      & 0.880 &  0.358 & 0.915 \\
\midrule   
\cite{learnN2learn} & {0.193}       & {0.815}      & {0.675} & {0.905} & {0.526} & {\textbf{0.924}} \\ 
\bottomrule
\end{tabular}%

\caption{\textbf{Comparison for sensitive attribute leakage:} We compare our approach on sensitive attribute leakage with the existing works. For the fairface dataset, sensitive attribute is race and task attribute is gender. In the CelebA dataset, sensitive attribute is gender and task attribute is smiling. The adversary accuracy is reported on the supervised reconstruction attack as described in~\ref{sec:attack}, for all the three methods, adversary accuracy is close to random chance, indicating that evaluation of privacy just by analyzing the adversary proxy during the training may give a false sense of privacy.}
\label{table:sensitive-attribute}
\end{table*}
\begin{table}[]
\centering
\resizebox{8cm}{!}{%
\begin{tabular}{l|l|l|l|l}
\toprule
& SSIM $\downarrow$ & PSNR $\downarrow$ & $\ell_1$ $\uparrow$ & Utility $\uparrow$ \\ \midrule
Traditional~\cite{osia2020hybrid} &  $0.88\pm0.03$ & $31.58\pm2.44$ & $108.82\pm8.92$ & \textbf{97.35} \\ \midrule 
Adversarial~\cite{learnN2learn}  &  $0.68\pm0.12$ & $20.49\pm5.94$ & $123.33\pm20.67$  & 97.15 \\ \midrule
DISCO   & \textbf{0.38$\pm$0.09}  & \textbf{11.61$\pm$1.91} & \textbf{125.34$\pm$15.29} & 95.66 \\ 
\bottomrule
\end{tabular}%
}
\vspace{-0.1cm}
\caption{\textbf{Comparison for sensitive input leakage:} We compare our approach on sensitive input reconstruction task and compare with our baselines and the existing works. \label{table:sensitive-input}}
\end{table}

%% file: tex/dataset_release.tex
\section{Benchmark for private inference}
As we strive towards rigorous understanding of privacy for collaborative inference, we also release an evaluation benchmark for attack models on sensitive inputs and attributes. The benchmark consists of 1 million pairs of activations, model weights, and inputs details for attacks discussed in this paper. The benchmark includes samples from 3 datasets: CelebA, CIFAR-10, and FairFace for multiple recent techniques focused on privacy during inference: DISCO, Max Entropy \cite{Roy_2019_CVPR}, and Adversarial \cite{learnN2learn}.

%% file: tex/results.tex

\begin{figure}
    \centering
    \includegraphics[width=.47\textwidth]{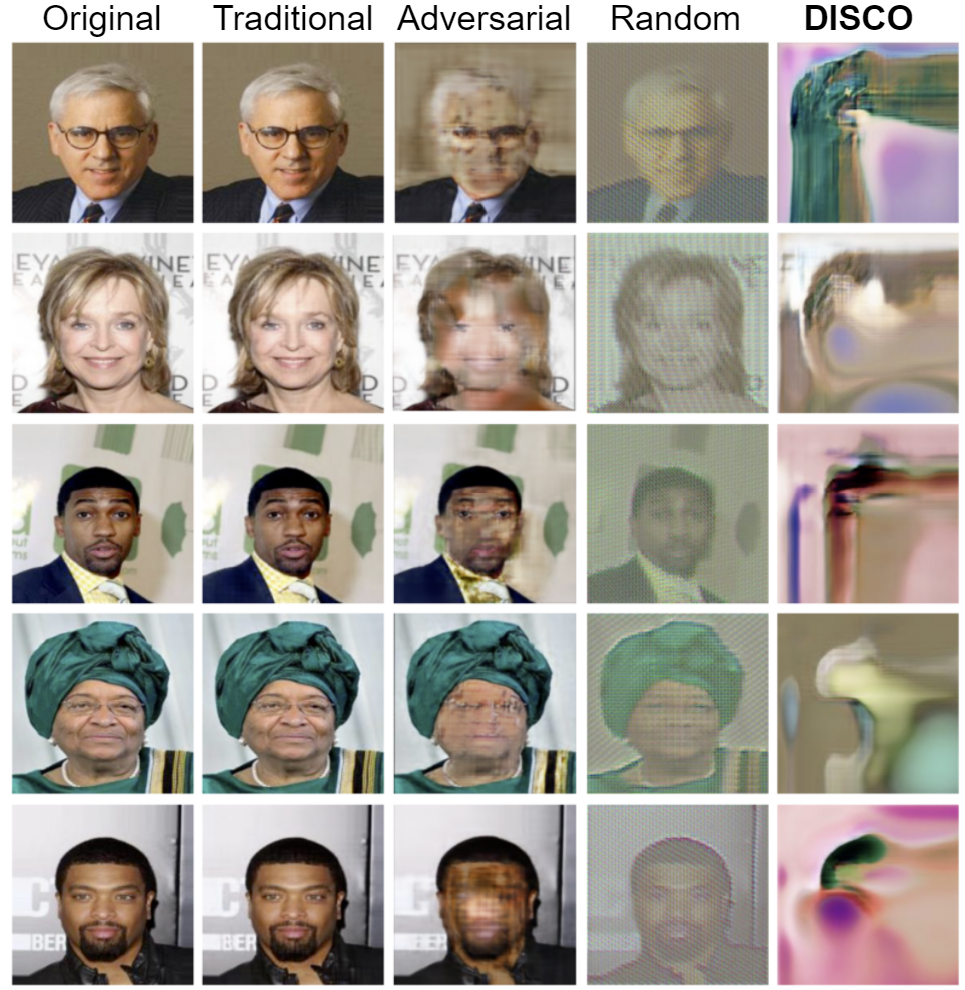}
    \caption{\textbf{Reconstruction attack qualitative evaluation}: We show the reconstruction quality across traditional collaborative inference, adversarial, and ours. Note that the utility performance on the target task of gender classification does not suffer from accuracy degradation in our task.\label{fig:quality-multifaces}}
    \vspace{-0.6cm}
\end{figure}

\section{Results}
\textbf{Sensitive Attribute.} For sensitive attributes, we perform qualitative analysis and report performance in Table~\ref{table:sensitive-attribute}. We mention accuracy of the adversary on the sensitive attribute (i.e. \textit{privacy}) and of the predictor on the task attribute (i.e. \textit{utility}). We note that DISCO provides the best \textit{privacy-utility} trade-off on each of these datasets. Specifically, on the CIFAR-10 dataset~\cite{krizhevsky2010convolutional}, without loss of utility, we improve on decreasing the adversary accuracy to \textbf{0.2282} from \textbf{0.3573} in ~\cite{Roy_2019_CVPR}, the most recent state-of-the-art. 

\textbf{Sensitive Input.} For sensitive inputs, we perform both quantitative (Table~\ref{table:sensitive-input}) and qualitative analysis (Figure~\ref{fig:quality-multifaces}) for stronger \textit{likelihood maximization} attacks. The visual results highlight that DISCO achieves significantly better obfuscation in the reconstructed input. This is corroborated by the quantitative results where DISCO obtains an SSIM of \textbf{0.38} and PNSR of \textbf{11.61} as against \textbf{0.68} and \textbf{20.49} for adversarial class of techniques~\cite{li2019deepobfuscator,learnN2learn}. Please note that while other techniques may also provide some level of obfuscation in reconstruction, DISCO is the \textit{only} technique which is able to additionally prohibit the \textit{re-identification} of the input image. (compare columns 3 and 5 in Figure~\ref{fig:quality-multifaces}). We observe that supervised decoder is a stronger attack for DISCO but loses the \textit{identity} of the original image. The results can be found in the supplementary.

Next, Figure~\ref{fig:reconstruction_fairface} presents the reconstructed outputs for varying depths of the client activations (from block-1 to block-7 of the ResNet-18~\cite{resnet}). The results indicate a progressive worsening of performance as we move towards shallower client activations (lower resnet blocks) for baselines. In contrast, we note that DISCO still consistently protects the input from the likelihood maximization attack. This validates the motivation and formulation of the \textit{pre-processing module} of DISCO. We evaluate privacy-utility tradeoff in the Figure~\ref{fig:tradeoff} by varying the trade-off parameter for both ARL~\cite{li2019deepobfuscator,learnN2learn} and DISCO.

\begin{figure}
    \centering
    \includegraphics{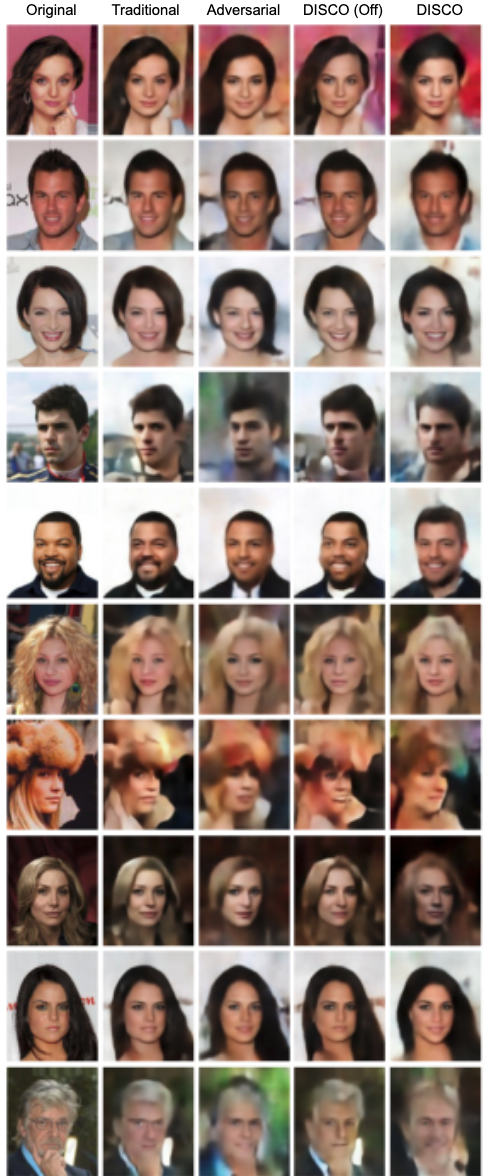}
    \caption{Qualitative comparison for different techniques using the supervised decoder attack described in the Section 3.1. \textit{DISCO (Off)} refers to DISCO with pre-processing module's toggle turned off. This technique results in a different yet realistic reconstruction for even \textit{DISCO} compared to deep image prior results shown in the Figure 3.}
    \label{fig:decoder_qual_results}
\end{figure}

We present more reconstruction results for the qualitative comparison in the Figure~\ref{fig:decoder_qual_results}. Our results indicate that supervised decoder based attack model performs significantly better than likelihood maximization attack for \textit{DISCO}, however, for all other techniques, likelihood maximization attack provides much better reconstruction quality. The figure can be found on the next page.

%% file: tex/discussion.tex
\section{Discussion}
In this section, we present the motivation and analyse the implication of various design choices for \textit{DISCO}.

\textbf{i) Privacy-Utility for Correlated Attributes}
While users idealize high privacy-utility guarantees, we posit that what level can be empirically realized is conditioned on the similarity of the task and sensitive attribute. To corroborate this position, we conduct leakage attacks using DISCO and \textit{traditional} with the following attribute configuration:
corroborate this with observations from the following experiments on the celebA dataset:
\begin{itemize}[nosep]
    \item \underline{\textbf{S1}}: Sensitive Attribute is \textit{Mouth Open} (yes/no) and the  Task Attribute is \textit{Smiling} (yes/no)
    \item \underline{\textbf{S2}}: Sensitive Attribute is \textit{Nose Size} and Task Attribute is \textit{Smiling} (yes/no)
\end{itemize}
Results in Table~\ref{tab:p-u-study} indicate DISCO achieves near-perfect privacy and high utility in S2, the privacy-utility worsens for S1 where the sensitive attribute (\textit{mouth open}) is strongly correlated with task attribute (\textit{smiling}) due to spatial overlap of the corresponding regions of interest. 
\begin{table}[t!]
\tabcolsep=0.1cm
\centering
\resizebox{7cm}{!}{%
\begin{tabular}{l|l|c|c}
\toprule
Sensitive Attribute              & Method      & Privacy ($\downarrow$) & Utility ($\uparrow$) \\
\midrule
\multirow{2}{*}{Mouth Open (S1)}   & ~\cite{osia2020hybrid} & 0.814                & 0.893             \\
                            & DISCO       & \bf{0.783}               & \textbf{0.907}              \\
\midrule
\multirow{2}{*}{Big Nose (S2)} & ~\cite{osia2020hybrid} & 0.616               & \textbf{0.896}              \\
                            & DISCO       & \bf{0.559}                & 0.893      \\
\bottomrule
\end{tabular}
}
\caption{Privacy-utility trade-offs is influenced by correlation of task and sensitive attribute. The task attribute here is \textit{Smiling} (yes/no). Both sensitive attributes are binary.\label{tab:p-u-study}}
\vspace{-0.2cm}
\end{table}

\textbf{ii) Comparing with Activation Noise for Privacy} Adding noise to the output of a statistical query (\textit{client activations} in this case) is a well known mechanism for privatizing sensitive data. These mechanisms are sometimes built under the framework of differential privacy~\cite{dwork2008differential} or its derivatives~\cite{mironov2017renyi,localDP}. While we do not compare or operate under a strict differentially private mechanism, we posit that preventing sensitive input reconstruction requires a heavy amount of noise. To validate this, we design an experiment where we add Gaussian noise to the \textit{client activations} and incrementally increase $\sigma$ until the reconstruction is prevented. We also measure the difference in utility obtained by these noise based mechanisms. Compared to the learning based approaches like adversarial and DISCO, achieving privacy through random noise comes at a heavy cost of deteriorating utility to the extent that utility gets close to random chance with noise that is empirically capable of preventing reconstruction attack $\mu=-1, \sigma=400$.

%% file: tex/conclusion.tex
\section{Conclusion}
In this work, we focus on selective privacy of sensitive information in learned representations. We posit that selectively removing features in this latent space can  protect  the  sensitive  information  and  provide  better privacy-utility trade-off. Consequently,  we introduce  DISCO,  a  dynamic  scheme  for  obfuscation of sensitive channels to protect sensitive information in collaborative inference.  DISCO provides a steerable  and  transferable  privacy-utility  trade-off  at inference without any retraining. We propose diverse attack schemes for sensitive inputs and attributes and achieve significant performance gain over existing methods on multiple datasets. To encourage rigorous exploration of attack schemes for private collaborative inference, we also release a benchmark dataset of 1 million sensitive representations.